\documentclass{article} 
\usepackage{collas2024_conference,times}
\usepackage{graphicx}
\usepackage{caption}
\usepackage{soul}

\usepackage{amsmath,amsfonts,bm}

\def\eqref#1{equation~\ref{#1}}

\def\1{\bm{1}}

\DeclareMathAlphabet{\mathsfit}{\encodingdefault}{\sfdefault}{m}{sl}
\SetMathAlphabet{\mathsfit}{bold}{\encodingdefault}{\sfdefault}{bx}{n}

\usepackage{hyperref}
\hypersetup{
    colorlinks=true,
    linkcolor=red,
    filecolor=magenta,
    urlcolor=blue,
    citecolor=purple,
    pdftitle={Overleaf Example},
    pdfpagemode=FullScreen,
    }

\title{iWISDM: Assessing instruction following in multimodal models at scale}

\author{
  \begin{minipage}[t]{0.45\textwidth}
    \raggedright\normalfont
    \textbf{Xiaoxuan Lei}\thanks{These authors contributed equally to this work.}\\
    Department of Physiology\\
    McGill University\\
    Mila, University of Montreal\\
    Canada\\
    \texttt{xiaoxuan.lei@mail.mcgill.ca}
  \end{minipage}
  \hfill
  \begin{minipage}[t]{0.45\textwidth}
    \raggedright\normalfont
    \textbf{Lucas Gomez}\footnotemark[1]\\
    Integrated Program in Neuroscience (IPN)\\
    McGill University\\
    Mila, University of Montreal\\
    Canada\\
    \texttt{lucas.gomez@mail.mcgill.ca}
  \end{minipage}
  \hfill
  \AND
  \begin{minipage}[t]{0.45\textwidth}
    \raggedright\normalfont
    \textbf{Hao Yuan Bai}\footnotemark[1]\\
    School of Computer Science\\
    McGill University\\
    Mila, University of Montreal\\
    Canada\\
    \texttt{hao.bai@mail.mcgill.ca}
  \end{minipage}
  \hfill
  \begin{minipage}[t]{0.45\textwidth}
    \raggedright\normalfont
    \textbf{Pouya Bashivan}\\
    Department of Physiology\\
    McGill University\\
    Mila, University of Montreal\\
    Canada\\
    \texttt{pouya.bashivan@mcgill.ca}
  \end{minipage}
}

\collasfinalcopy 

\begin{document}

\maketitle

\begin{abstract}


The ability to perform complex tasks from detailed instructions is a key to many remarkable achievements of our species. 
As humans, we are not only capable of performing a wide variety of tasks but also very complex ones that may entail hundreds or thousands of steps to complete. Large language models and their more recent multimodal counterparts that integrate textual and visual inputs have achieved unprecedented success in performing complex tasks. Yet, most existing benchmarks are largely confined to single-modality inputs (either text or vision), narrowing the scope of multimodal assessments, particularly for instruction-following in multimodal contexts. To bridge this gap, we introduce the instructed-Virtual VISual Decision Making (iWISDM) environment engineered to generate a limitless array of vision-language tasks of varying complexity. Using iWISDM, we compiled three distinct benchmarks of instruction following visual tasks across varying complexity levels and evaluated several newly developed multimodal models on these benchmarks. Our findings establish iWISDM as a robust benchmark for assessing the instructional adherence of both existing and emergent multimodal models and highlight a large gap between these models' ability to precisely follow instructions with that of humans. Code is available at \url{https://github.com/BashivanLab/iWISDM}.

\end{abstract}

\section{Introduction}
\begin{figure}[th]

\begin{center}
\includegraphics[width=1.0\textwidth]{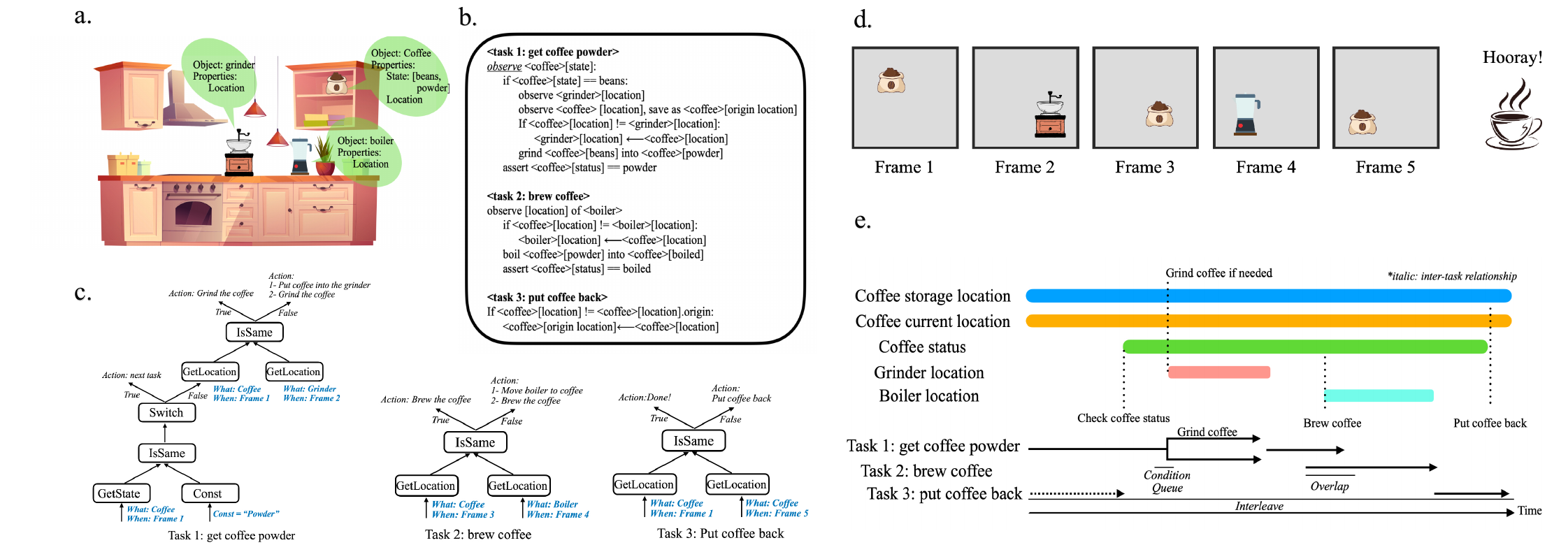}
\end{center}
\caption{Illustration of coffee making task as an example real world compositionally constructed task. \textbf{a)} Cartoon depiction of a kitchen with typical objects within. The dialogue boxes highlight the relevant properties of the grinder, coffee bag, and boiler; \textbf{b)} Pseudocode detailing the coffee-making task, encompassing three subtasks – obtaining coffee powder, brewing the coffee, and returning the coffee; \textbf{c)} Graphical representation of the three subtasks as computational graphs. Blue text highlights the properties associated with each operator; \textbf{d)} An abstraction of the real-life coffee-making task as a task in iWISDM. Frame sequences are generated using the task graphs illustrated in \textbf{c}; \textbf{e)} Graphic depiction of object properties actively in use (top) and subtasks (bottom). Coloured bars in the upper panel signify the persistence of each object property during the execution of coffee-making task. Lower panel depicts the interrelations between subtasks and various temporal compositional operations.
}
\label{fig:1}
\end{figure}

A typical day in most people's lives involves hundreds or thousands of tasks. Most of which are performed without explicit attention. Just in between getting up and getting to work, one may have already performed 5-15 tasks (taking a shower, shaving, making coffee, getting dressed, etc.). Teaching artificial agents to perform similar seemingly mundane tasks has proven to be an extremely difficult computational problem \citep{konar2018artificial}. The challenge becomes more apparent when one realizes that each of these seemingly mundane tasks such as making coffee involves tens of steps/actions (Figure \ref{fig:1}). 
The challenge becomes increasingly more significant once we consider more complex tasks such as operating a device or assembling a piece of furniture from its instruction manual. And yet, these tasks are performed proficiently by most individuals in most situations.  

Large Language Models (LLMs) have become increasingly capable of comprehending natural language across wide topics and contexts, allowing them to hold meaningful conversations, give expert advice, and analyze data among other features \citep{brown2020language, ouyang2022training, radford2019language}. In the meantime, their multimodal counterparts are starting to emerge, signalling broader application of such models across industries. Large Multimodal Models (LMMs) are generally capable of receiving and responding in a range of possible modalities including visual, text, and audio \citep{alayrac2022flamingo, liu2023visual, achiam2023gpt}. For example, the Gemini-Ultra model, which accepts text, image, audio, and video inputs, and responds with a combination of text and image outputs, recently achieved state-of-the-art on a range of single and multimodal benchmarks \citep{team2023gemini}.

However, existing benchmarks for assessing such models face several shortcomings: 
\textbf{(1)} Most multimodal benchmarks like FLEURS (audio-based, \citep{conneau2023fleurs}) and VATEX (video-based, \citep{wang2019vatex}) are still unimodal in their inputs, and do not permit detailed assessment of models' capacity to integrate information across modalities towards task goals. \textbf{(2)} Visual Question-Answering (VQA) datasets like VQAv2 \citep{goyal2017making} and CLEVR \citep{johnson2017clevr} assess reasoning with visual information in static images without addressing temporal information integration and sequential decision-making. \textbf{(3)} Open-ended learning environments such as XLand \citep{team2021open}, Crafter \citep{hafner2021benchmarking}, and Minecraft \citep{guss2019minerl} have been utilized for training reinforcement learning agents. It remains unclear whether and how they can be adapted to benchmark LMMs. \textbf{(4)} To the best of our knowledge, none of the existing benchmarks specifically assess models' ability to precisely follow instructions in the context of decision making tasks, an important measure of reliability and trustworthiness. Despite its importance, conducting such assessments has been particularly challenging because of the difficulty of collecting samples of multi-step tasks with ground truth information. \textbf{(5)} More recent benchmarks such as MME \citep{fu2023mme}, Mmbench \citep{liu2023mmbench} and MMvet \citep{yu2023mm} cover a wide range of cognitive tasks that start to adopt manually-generated or GPT-powered responses. However, those benchmarks are difficult to scale, which makes them inconvenient when investigating the scaling properties of LMMs. In addition, benchmarks such as OwlEval \citep{ye2023mplug} and LVLMeHub \citep{xu2023lvlm} rely on subjective human responses that often show high variability across individuals.



To close this gap, we designed \textbf{i}nstructed-\textbf{V}irtual \textbf{V}i\textbf{s}ual \textbf{D}ecision \textbf{M}aking (iWISDM), a virtual environment that enables procedural generation of complex, multi-step decision making tasks that test an agent's capacity to process visual information guided by natural language instructions. iWISDM builds on the compositional nature of natural behaviour \citep{barker1963stream} and the fact that complex tasks are often compositionally constructed by combining smaller task units in time. We thus developed a framework which allows instantiating visual decision-making tasks as computational graphs that could be combined logically and temporally to construct a virtually infinite number of tasks with varying complexity. The code of iWISDM is available on GitHub at \url{https://github.com/BashivanLab/iWISDM}.

Our main contributions are: 
\begin{enumerate}
    \item We introduce iWISDM, a virtual environment for the procedural generation of limitless visual decision-making tasks accompanied by natural language instructions. 
    \item We use iWISDM to construct three vision-language multimodal benchmarks with varying complexity levels to probe LMMs' ability to follow natural language instructions. 
    \item We test several recently developed LMMs as well as human subjects on these benchmarks and identify a notable shared weakness across existing LMMs compared to humans in their ability to precisely follow user instructions in the context of visual decision-making tasks. 
\end{enumerate}

\section{Related Works}
\subsection{Large Multimodal Models}

Continual advancements in pretrained large-scale multimodal models are driving progress in a wide array of downstream tasks. Given the significant computational expense associated with end-to-end pretraining, it is common to utilize readily available pretrained vision models alongside Large Language Models (LLMs) such as (OPT \citep{zhang2022opt}, FlanT5 \citep{chung2022scaling}, Vicuna \citep{chiang2023vicuna}, LLaMA \citep{touvron2023llama}). 
Pioneering LMMs such as VisualGPT \citep{chen2022visualgpt} and Frozen \citep{tsimpoukelli2021multimodal} have highlighted the advantages of leveraging pre-trained multimodal models. The primary challenge lies in achieving cross-modal alignment, given that LLMs typically lack exposure to images during their unimodal pretraining phase. LMM research is coalescing around a key strategy known as the \emph{``visual instruction tuning"}, which involves a two-phase training process: firstly, a vision-language alignment pretraining stage, and secondly, a visual instruction tuning stage. 

A range of methods and models have been developed to enhance the capabilities of LMMs. Early approaches use a frozen object detector for visual feature extraction \citep{chen2020uniter, li2020oscar, zhang2021vinvl} while LiT \citep{zhai2022lit} borrowed a frozen pretrained image encoder from CLIP \citep{radford2021learning}. More recently, Frozen \citep{tsimpoukelli2021multimodal} and Flamingo \citep{alayrac2022flamingo} have adopted an image-to-text generation approach, prompting the language model to generate text based on an input image. However, in BLIP-2 \citep{li2023blip}, it was shown that this approach is not adequate for overcoming the modality gap and instead proposed a Querying Transformer (Q-Former), acting as a visual resampler, and a two-stage bootstrapping pretraining method, which led to models that outperformed Flamingo80B \citep{alayrac2022flamingo} in zero-shot VQAv2 tasks with fewer trainable parameters. The Q-former architecture was also adopted by later works such as InstructBLIP \citep{dai2023instructblip} and Qwen-VL \citep{bai2023qwen}. 

Moreover, GPT-4 \citep{achiam2023gpt} has demonstrated remarkable proficiency in multi-modal dialogues with humans. Models like LLaVA \citep{liu2023visual} and MiniGPT-4 \citep{zhu2023minigpt} have sought to emulate its performance by integrating a fully connected vision-language cross-modal connector, which significantly reduces the need for paired image-text data during pretraining. Both have shown notable proficiency in following natural instructions and visual reasoning. 
Beyond image-based LMMs, there have been developments in models that specialize in video information processing. PaLM-E \citep{driess2023palm} incorporates continuous real-world sensor data into LMMs, facilitating the unification of real-world perceptions with human language. Video ChatCaptioner \citep{chen2023video} leverages ChatGPT’s conversational interface to enhance its understanding of video spatiotemporal contexts. 
\subsection{Multimodal benchmarks}
The rapid progression of LMMs has driven the need for comprehensive benchmarks to assess their multifaceted capabilities. Traditionally, 
datasets primarily focus on computer vision tasks -- classification, detection, segmentation, captioning, visual generation, and editing \citep{CVinW_Readings} -- where instructions are implicitly integrated. However, these primarily unimodal datasets do not adequately test the models' proficiency in multimodal information alignment, revealing a limitation in our ability to fully evaluate the LMMs. 

To address this, the community has turned to human-annotated datasets like MS-COCO \citep{lin2014microsoft} and web-scraped collections such as YFCC-100M \citep{goyal2019scaling} to forge better-aligned multimodal datasets. Enhanced by additional data cleansing methods and supported by CLIP-like models, datasets like Conceptual 3M/12M \citep{lai2023scarcity} and LAION-5B \citep{schuhmann2022laion} have not only improved in descriptive quality and text-image alignment but also in scale. 
Despite these advancements, a crucial capacity of LMMs which is to effectively follow multimodal vision-language instructions, remains inadequately assessed by current benchmarks. To address this gap, several benchmarks have pivoted toward evaluating cognitive skills and systematic, quantitative assessments. Visual Question Answering (VQA) datasets like ScienceQA \citep{saikh2022scienceqa} play a crucial role in examining LMMs' multimodal reasoning capabilities. More recent benchmarks such as MME \citep{fu2023mme} and LLaVA-Bench \citep{liu2023improved} utilize manually crafted questions and answers for precise evaluations, while platforms like MMBench \citep{liu2023mmbench} adopt ChatGPT-driven techniques for data creation and response generation. Notably, ShareGPT4V \citep{chen2023sharegpt4v} introduces a dataset with high-quality captions, initially derived from GPT4-Vision and subsequently expanded, reflecting a trend towards more sophisticated and scalable evaluation frameworks.



The exploration into video-based datasets expands the assessment scope to include temporal integration, covering areas like video captioning, event segmentation, and action prediction. Datasets from video game environments (e.g., Minecraft \citep{guss2019minerl}, XLand \citep{team2021open}, and Crafter \citep{hafner2021benchmarking}) and Video Question Answering tasks (e.g., YouCook2 \citep{zhou2018towards}) are pivotal in evaluating models’ strategic understanding and instructional adherence. Additionally, benchmarks like Seed-Bench-2 \citep{li2023seed} and various perception tests pose further challenges for LMMs by testing their efficacy in navigating and interpreting complex, multimodal data streams. Yet, a gap persists in evaluating models' precision in following instructions across sequential images, a gap that the iWISDM environment aims to bridge.


\section{Methodology}
We developed iWISDM to facilitate the generation of a diverse range of sequential visual reasoning tasks, which vary in complexity and require minimal user intervention. iWISDM encompasses a broad spectrum of tasks that engage executive functions such as inhibition of action, working memory, attentional set, task switching, and schema generalization. These functions are traditionally associated with the prefrontal cortex, a critical area for advanced cognitive processes in the brain \citep{fuster2015prefrontal,sun2023organizing}. Notably, the iWISDM task space is designed to accommodate classic working memory and decision-making tasks commonly employed in neuroscience and cognitive science research \citep{rigotti2013importance, goldman1992working, fuster2009cortex}. In the following sections, we begin by detailing the key functionalities of the iWISDM environment, followed by an in-depth description of its design and the implementation of its constituent components.

\subsection{Design}
\label{sec_design}
Inspired by prior work \citep{yang2018dataset}, iWISDM generates tasks through a 3-phase procedure: \textbf{(1)} Task graph construction; \textbf{(2)} Node initialization \textbf{(3)} Trial instantiation. Dividing the generation procedure into distinct phases eliminates the necessity of constructing the task graph anew for each task trial. Once all properties associated with the nodes in a given task graph have been specified (phase 2), the user can generate any number of trials of that particular task, each with potentially different stimuli, ground-truth actions, as well as any other inherent stochastic values such as the number of delay frames. In general, iWISDM creates tasks following: 

$$\textbf{f}, i, \textbf{r} = \texttt{iWISDM}(G)$$

where, \textit{G} denotes the task graph, \textbf{f} the sequence of visual frames, \textit{i} the corresponding language instructions, and \textbf{r} the sequence of ground-truth actions for each visual frame within \textbf{f} according to the instruction \textit{i}. 

The task graph \textit{G} can either be specified by the user or generated automatically via \texttt{AutoTask}. The major distinction between the two resides in the initial task graph construction phase (phase 1). In contrast to the user-specified mode, where the user needs to manually define the task graph, in \texttt{AutoTask} mode, the user needs to only specify the parameters listed in the following section. 

\subsubsection{Task graph construction}
In iWISDM, nodes and edges create directed, acyclic, connected task graphs. Each node represents a predefined operator that contributes to defining the task (each node in an iWISDM graph represents a task operator, so we use the terms 'node' and 'operator' interchangeably). Task operators take downstream stimuli/actions as input and output stimuli/actions based on their definitions. While some operators must have parent/upstream operators, root operators define the actions of a task. They form minimal sub-graphs that define sub-tasks (e.g. \texttt{Get} operators are root operators that define the subtask: \textit{What is the attribute of an object?}). Under user-defined connectivity rules, sub-graphs can be combined to generate corresponding compositional tasks. In our environment, the \emph{depth} of the graph is measured by the longest path from the root operator to any other operator within the graph.

Each operator has a customizable set of rules that constrain its connections. The specific operators and their permissible connections are described below:  (see Figure \ref{fig:3} for visualization):
\begin{itemize}
    \item \textbf{Functional Operators:} \texttt{Select}: This operator defines stimuli based on three criteria: when, where, and what. ``When'' refers to the specific frame from which the stimuli originate. ``Where'' indicates the location of the stimuli within the frame. ``What'' depends on the particular dataset from which the stimuli are derived. For example, in our ShapeNet environment, the stimuli have three attributes: category (such as car versus plane), identity (which specific car), and view angle (the angle from which the stimulus is rendered). \texttt{Select} operators that have no downstream connections are the terminal nodes of the graph. Conversely, its potential downstream operators may include any \texttt{Get*} functional operator. \texttt{Switch}: Based on the output action of a boolean task, this operator connects the logic to one of two possible paths. Its compulsory downstream connection must be a boolean operator, while its typical upstream connections are subtasks graphs. \texttt{Get*}: This group of operators is responsible for fetching specific properties of stimuli, such as category, location, or identity, exemplified by operators like \texttt{GetCategory}, \texttt{GetLocation}, and \texttt{GetIdentity}. Its direct downstream connection is always a \texttt{Select} operator and its upstream connections can be any boolean operator. \texttt{CONST}: The simplest operator, \texttt{CONST} represents a fixed value. It is often used as a downstream connection for boolean operators that compare attributes.
    \item \textbf{Boolean Operators:} \texttt{Exist:} Paired with a specific property value (e.g., 'Desk'), this operator poses the question of whether an object with the property ('Desk') exists. It generates a boolean output and functions as an action generator or downstream connection for boolean operators. \texttt{And}, \texttt{IsSame}, \texttt{NotSame}, \texttt{Or}: These boolean operators process inputs from two boolean operators to produce a boolean outcome. They are critical in constructing logical conditions within the task graph.
\end{itemize}

With these operators, iWISDM provides two modes for constructing task graphs: user-specified and automatic. 
\begin{itemize}
    \item \textbf{User-specified task graph construction.} In this mode, users have the freedom to fully specify the task graph by manually creating an instance of NetworkX directed graph. Doing so requires a manual definition of all nodes (operators) and edges (connection between operators) within a task graph, thereby establishing the desired task operation logic. Subsequently, the task graph can be fed to the generator function (\texttt{write\_trial\_instance}) to yield the corresponding task trials. 
    One potential application of this mode is to replicate trials from a specific task such as the \emph{n-back} or \emph{contextual decision-making} tasks. In our core build of iWISDM, we have incorporated a collection of classic tasks from neuroscience and cognitive science literature that users can readily access.
    
    \item \texttt{AutoTask}\textbf{ graph construction.} In \texttt{AutoTask} mode, users can define a custom \emph{task space} with a set of hyperparameters to procedurally generate tasks. A task space delineates the complexity and permissible operations pertinent to task construction. The available hyperparameters are: \textbf{(1)} number of compositions, i.e. maximum number of \texttt{Switch} operators to compose subtasks \textbf{(2)} each task graph's maximum depth and maximum number of operators \textbf{(3)} set of operators to sample from. 
\end{itemize}
    
In addition to the above hyperparameters, in \texttt{AutoTask} mode, the allowed task structure is further constrained by the permitted connectivity for various operators (e.g. \texttt{And} operators must be followed by other boolean operators such as \texttt{IsSame}, \texttt{NotSame}, \texttt{And}, \texttt{Or}). A default operator connectivity is defined for all existing operators in our core build, but new connectivity rules could easily be added for any new user-defined operators. This is done through a Python dictionary, which details the allowed input and output operators for additional operators. By specifying these hyperparameters, iWISDM autonomously generates random runnable task graphs derived from the predefined task space. Each resultant task graph can then be used to generate specific trials. 

To ensure each task graph would comply with the connectivity rules between operators, we follow a backward initialization process during \texttt{AutoTask} (Figure \ref{fig:S1}). The task generation process starts from the root node and descends recursively. For each current node/operator $n$ in the graph, its downstream operators $C_n$ are randomly sampled based on the connectivity rules. As the graph depth approaches the specified maximum depth, the permissible operators with the shortest possible subtask depth are sampled into $C_n$. For instance in our core build, if $n$ is the \texttt{And} operator, then only \texttt{IsSame}, \texttt{NotSame} are sampled since they have shorter subgraph depths than \texttt{And}, \texttt{Or}. Through this procedure, iWISDM \texttt{AutoTask} facilitates sampling from diverse task spaces with varying degrees of complexity specified by the user. The utilization of the connectivity rule dictionary and \texttt{Switch} operator guarantees the logical and feasible nature of the generated tasks, providing researchers with an extensive pool of tasks for investigation and exploration.

Together, iWISDM’s two operating modes provide users with the flexibility to use the environment to either train or evaluate models on specific tasks (e.g. classic tasks from literature), as well as a wide variety of tasks adhering to specific guidelines as stipulated by the specified hyperparameters.

\subsubsection{Node Initialization}
The second step assigns values to each node within the task graph, thereby conferring logically coherent tasks. There are two critical challenges: the initialization of an independent task graph and the integration of multiple graphs in time.

To instantiate a task graph (Figure \ref{fig:S1}a), a backward recursive approach is used \citep{yang2018dataset}, similar to that of \texttt{AutoTask} task graph generation. During trial instantiation, given the expected stimuli/action output, an operator propagates stimuli-related properties or actions to its children operators in reverse topological order. The process starts from the root operator of the graph that corresponds to the final task action. To guarantee a balanced action space, we uniformly sample from the pool of possible output (e.g. true/false, location values, and object categories) and assign it to each unassigned operator. We then iteratively go one layer down until reaching the \texttt{Select} leaf nodes, propagating the expected output to all nodes. At this stage, each task only has one output action. 


The backward process is confined to the instantiation of individual task graphs and ensures logical consistency within each task. To ensure inter-graph logical consistency, we formulated a forward algorithm, generating complex tasks that require output actions at different frames, which we call temporal composition (see section \ref{main_features}. During this process, distinct \texttt{Select} operators might assign conflicting attributes to the same stimulus. The forward process (Figure \ref{fig:S1}b) therefore also serves to resolve disparities between \texttt{Select} operators in the same frame. It identifies the earliest merging frame, and resolves potential conflicts on a frame-by-frame basis, adjusting the properties of ensuing operators.

In summary, the node initialization process focuses on property assignment, involving the initialization of individual graphs followed by fusing them together. This process aims to yield logically coherent task instances, both within subtasks and in temporally merged tasks.
\begin{figure}[th]
\centering
\includegraphics[width=1.0\textwidth]{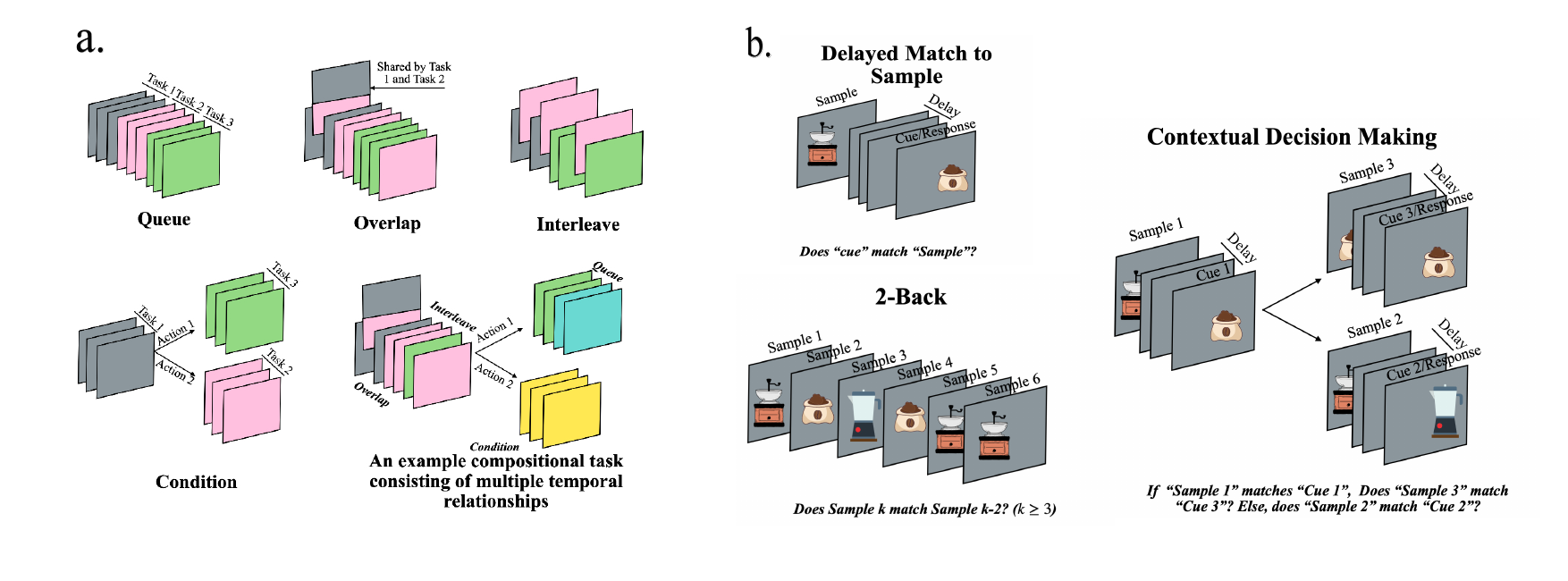}
\caption{Illustration of various temporal compositional structures used in iWISDM and their application in instantiating classic decision-making tasks. \textbf{a)} Four distinct temporal composition operations including \textit{Queue}: where one task is completed before starting the next; \textit{Overlap}: where two or more tasks share common information; \textit{Interleave}: depicting the interwoven acquisition of information related to different tasks and; \textit{Condition}: where the execution of a subsequent task depends on the outcome of the preceding one. An example compositional task consisting of multiple temporal relationships is shown. Frames are coloured differently to highlight distinct tasks, with multicoloured frames indicating shared information across tasks; \textbf{b)} Three classic cognitive tasks are exemplified: Delayed Match to Sample; 
2-Back; 
and Contextual Decision Making. 
}
\label{fig:2}
\end{figure}

\subsubsection{Task trial instantiation}
Regardless of the selected operation mode, for each task trial, iWISDM yields a frame sequence, an accompanying natural language instruction, and an action sequence. As delineated above, tasks are initially formulated with graphs that define the task logic. The task graph then serves as the basis for instantiating task trials. Distractors and fixation cues are also added during this step. Each task trial comprises of the following distinct components:

\begin{itemize}
    \item \textbf{Frame sequence}. The frame sequence consists of an array of images. They display the visual information at each time step of the trial. Each frame is accompanied by a dictionary that contains the object properties within that frame. Images are stored as \texttt{PNG} files. 
    
    \item \textbf{Natural language instructions}. Each trial is accompanied by a natural language instruction that explains the task steps and decision criteria. Natural language instructions are generated concurrently during task instantiation. A partial string is assigned to each operator in the task graph that depends on its definitions and initialization (see Figure \ref{fig:S1} for an example). This approach allows iWISDM to automatically produce contextually relevant natural language instructions that describe the task in a human-readable format for each task. 
    \item \textbf{Action sequences}. Each trial also consists of an array of ground-truth actions at each frame of the trial. The action sequences could be used for supervised training and validation of the agents on the generated trials. 
     \item \textbf{Distractors (optional)} For users interested in making trials more attention-demanding, we offer a pre-implemented solution that adds distractors post-hoc. By specifying parameters during trial generation, distractors can be added without causing conflicts with existing task rules. The major challenge lies in distinguishing between stimuli and distractors in task instructions. This is achieved by specifying the attributes of the \textit{stimulus} that are not used during task execution. Additionally, it is important to avoid using distractors that share the same non-used attribute values. Based on the instructions, an ideal agent should be able to identify the stimulus of interest accurately. An example can be seen in Appendix Figure \ref{fig:S4-distractors}. 
 
\end{itemize}





\begin{figure}[h]
\centering
\includegraphics[width=.9\textwidth]{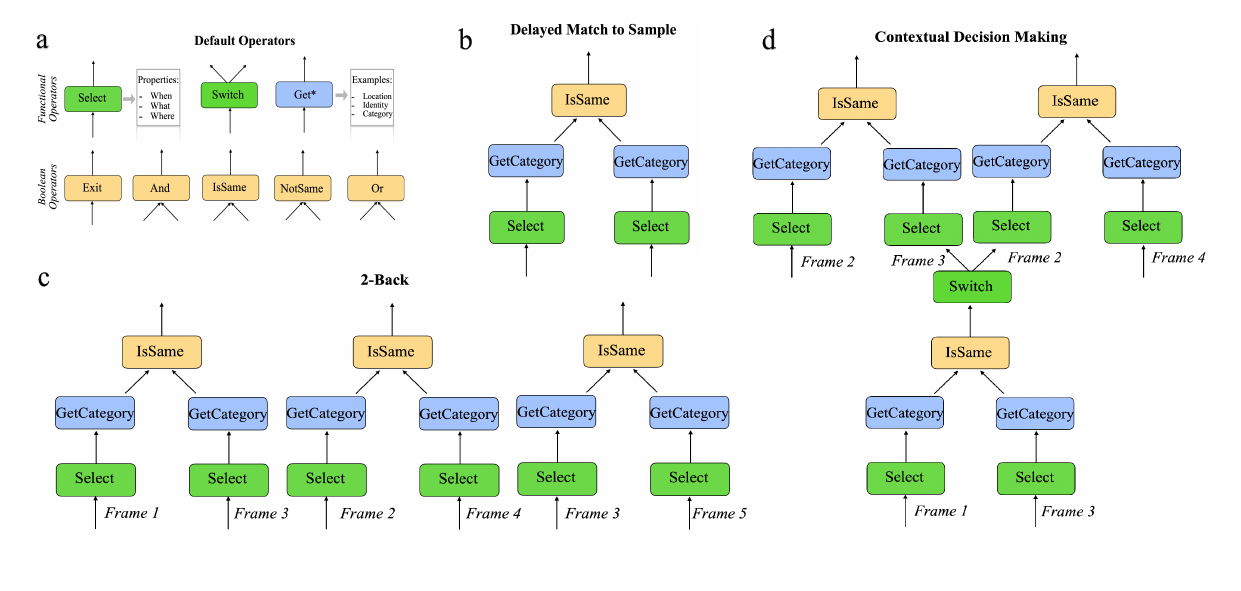}
\caption{Description of operators and task graph examples in the core build of iWISDM. \textbf{a)} Two main types of operators are considered. Functional Operators: \texttt{Select} operators retrieve stimuli according to specified attributes such as time and location; \texttt{Switch} operators accept boolean values and direct the task logic to one of two different subtasks; \texttt{GetAttribute} operators take stimuli as input and output the corresponding attribute values of those stimuli. Boolean Operators comprise \texttt{Exit}, \texttt{And}, \texttt{Or}, \texttt{IsSame}, and \texttt{NotSame}. These operators take boolean values as inputs and produce another boolean value based on their boolean logic. \textbf{b-d)} Using the operators defined in panel \textbf{a}, we create tasks from pre-specified rules. Panels \textbf{b}, \textbf{c}, and \textbf{d} display the task graphs for the three typical cognitive tasks illustrated in Figure \ref{fig:2}b, demonstrating how these operators are interconnected to define the structures of these specific tasks. For instance, the task instruction for panel \textbf{b} and \textbf{d} are: \textit{``category of object 1 equals category of object 2?''}, \textit{``if category of object 1 equals category of object 3, then category of object 2 equals category of object 3, else category of object 2 equals category of object 4?'}. The task instruction for panel \textbf{c} is similar to panel \textbf{b} but repeated across time. 
}
\label{fig:3}
\end{figure}

\subsection{Main Features}\label{main_features}
At its core, iWISDM is designed to be scalable and extensible. To do so, we adopted a modular framework in which task rules are constructed compositionally by combining functional and boolean operators. The combination of these operators gives rise to distinct tasks. Likewise, \emph{task spaces} (i.e. collections of instantiable task graphs) are spanned by specifying the set of allowable operators and operator-operator connection rules (see section \ref{sec_design} for detailed definitions). We detail iWISDM's main features below.\\

\textbf{Compositionality}. Real-world tasks are fundamentally compositional, as most tasks can be readily decomposed into sets of simpler subtasks that involve fewer sensory observations and cognitive operations. Two crucial facets of compositionality need to be considered: logical and temporal, both of which are common in daily human behaviour and allow individuals to efficiently handle complex tasks and adapt to dynamic environments. Cognitive processes involving task decomposition (i.e. breaking tasks into subtasks) and temporal combination of decision rules (i.e. combining the outcomes of subtasks) are fundamental to our ability to navigate the world around us.

\begin{itemize}
  \item \textbf{Logical}: An agent's action can be viewed as a function of sensory observations, internalized world knowledge, prior actions, and its objectives \citep{ha2018world}. Yet, as outlined in the introduction, the decision-making process frequently decomposes into sub-decisions and information-processing steps that are temporally constrained and require fewer observations.
  We define \emph{logical compositionality} as how decisions can be combined hierarchically through boolean operators (i.e. \texttt{And}, \texttt{Or}, etc) and functional operators (i.e. \texttt{Switch} operator that asks for if...then...). As an example, consider the contextual decision-making task (ctxDM, see Figure \ref{fig:2}b or Figure \ref{fig:3}d). In the task in Figure \ref{fig:3}d, the subject needs to first compare the category of the objects in the first and third frames. If their categories are the same, then compare the category of the objects from the second and \textit{third} frames. Otherwise, compare the category of the objects from the second and \textit{fourth} frames. For this task, the task rule of the second task is conditioned upon the result of the first task, which makes a logical composition. As the depth and the total number of operators involved in the task grows, we can compose more complex logical structures.

  \item \textbf{Temporal}: Temporal compositionality is concerned with how different decision rules should be combined together to construct a complex task that extends \emph{in time}. In real-world scenarios, individuals often face tasks that require multiple decisions to be made in sequence or in parallel. For instance, making coffee involves following decision rules to accomplish a sequence of tasks such as grinding the coffee beans, brewing, and pouring (Figure \ref{fig:2}). While rule compositionality has been explored in several prior works \citep{lake2023human, livska2018memorize, loula2018rearranging}, the topic of temporal compositionality has attracted less attention in the field. This is potentially due to a lack of proper datasets or virtual environments that could enable such investigations. iWISDM is precisely engineered to fill this gap by enabling the generation of temporally compositional tasks made from combining simpler tasks in time (Figure \ref{fig:2}). 
  
\end{itemize}

\textbf{Vast Task Space}. Another important feature of iWISDM is the vastness of its task space. This feature naturally extends from the compositionality that is inherent to iWISDM's task generation procedure. The ability to produce a large number of distinct tasks is critically important for the robust evaluation of large multimodal models that are trained on increasing volumes of information from the web.
Moreover, the vast number of instantiable tasks in iWISDM provides an opportunity for training or fine-tuning large multimodal models to improve their ability to follow instructions in a vision-language context. 

\textbf{Natural Language Task Instruction}. Natural language provides a rich and convenient way of communicating complex information to biological or artificial agents. It has been shown that improvements on language understanding in LLMs directly enhance their generality (performing many tasks) and adaptation (0-shot generalization)\cite{brown2020language, radford2019language}. Due to their capacity to compress enormous knowledge bases, these models have been useful in various applications where traditionally human supervision had been necessary \cite{shah2023lm, dasgupta2023collaborating}. Perhaps for similar reasons, natural language input constitutes the core of most existing multimodal models and they are heavily trained on large text corpora among others.
For this reason, in iWISDM, each task is accompanied by a simplified natural language instruction (see examples \ref{fig:S1}). When completing complex tasks, the instruction communicates first the task structure in terms of upcoming observations, then the task rules that determine the relationship between observations and actions. 

\textbf{Automatic Task Generation}. In contrast to prior virtual environments tailored for cognitive or neuroscience investigations \citep{molano2022neurogym}, iWISDM can not only generate hand-crafted cognitive tasks such as classical decision-making tasks (e.g. contextual decision-making and n-back, \ref{fig:3}), but also allow procedural task generation from a pre-specified task space defined by a small set of hyperparameters (See Section 3.2 for details). 

\textbf{Customizability and Extensibility}. We envision the future of iWISDM as a framework that will be continuously developed and expanded by the larger community of machine learning scientists, cognitive scientists, and neuroscientists. For this reason, we have designed iWISDM to be highly customizable and extendable in task operators, visual inputs, and stimulus properties (for detailed discussions, see Appendix \ref{appendix_extensible}).

\section{Evaluation}
\subsection{Models \& Humans}
As a preliminary evaluation, we test the capabilities of GPT-4V, Gemini-Pro-1.0, Claude-3, InternLM-XComposer2, and MMICL in solving iWISDM tasks\footnote{All models were evaluated between the dates of January 24th, 2024 and April 5th, 2024.}\footnote{All models were set to generate a maximum of between 5 and 8 tokens depending on the required task response type.}. We compare their performance to human baselines and find a notable gap in the multi-image instruction-following task capabilities of existing LMMs. 

We also collected responses from 6 human subjects tasked to answer three sets of randomly selected trials (Figure \ref{fig:Ss},\ref{fig:S3}, \ref{fig:S4}), each set sampled from a different complexity level (total of 150 trials). The task trials were displayed in a way similar to that of the models, where images can be seen alongside the task's text instruction following a general task description.

\subsection{Complexity}
Using the \texttt{AutoTask} framework in iWISDM, we created three benchmarks corresponding to Low, Medium, and High complexity tasks. Low complexity tasks were restricted to contain exactly one logical joining operator ("and/or"), exclude \texttt{Switch} operators, and require only boolean actions. Medium complexity tasks had the same logical joining operator restriction as low complexity and boolean action requirements, with the complexity increase coming from adding a \texttt{Switch} operator. Finally, high complexity tasks have between one and two logical joining operators, an included \texttt{Switch} operator, and require boolean and object property action responses (i.e. '... category of object 1?', Answer: 'planes'). For further details refer to Figure \ref{fig:complexity-table}. Additionally, since the benchmarks are generated based on a synthetic stimuli dataset, while all LMMs are trained with naturalistic stimuli, we aim to confirm whether LMMs can recognize rendered ShapeNet objects. Therefore, we developed two sets of simplistic single-frame tasks: location-only and category-only existence tasks (e.g. Is object 1 in the bottom left?). Evaluation on these two single-frame task sets provides an upper bound of LMMs' performance.

The set of tested models was limited due to the scarcity of applicable models. Many popular open-source LMMs, such as LLaVa-1.5, were unable to perform the task simply due to their limited image sequence lengths. A minimum image sequence length of ten is needed to complete all complexity levels. We were able to properly evaluate two open-source models, InternLM-XComposer2-7b and MMICL-Instructblip-T5-xl. For samples of prompts used to evaluate each model see Appendix \ref{fig:prompts}. 

\subsection{Results}

\begin{figure}[h]
\centering
\vspace{-0in}
\includegraphics[width=1\textwidth]{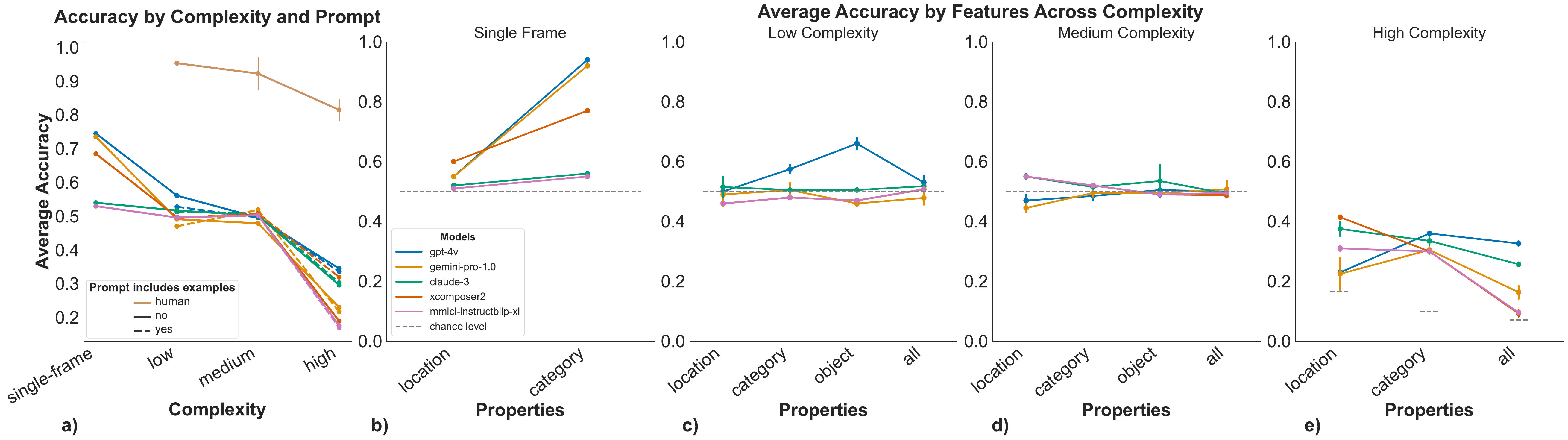}
\caption{\textbf{a)} Accuracy of LMMs and Humans on iWISDM benchmarks of varying complexity, and prompt type. \textbf{b), c), d), \& e)} Average accuracy of applicable models on single-frame, low, medium, and high complexity tasks for each feature type. Chance level represents the baseline accuracy of randomly guessing one of the applicable actions.} 
\label{fig:4}
\end{figure}

Figure \ref{fig:4}a shows the accuracy of actions taken by the models plotted against the complexity level for each type of prompt. GPT-4V generally achieves the best model scores, with the largest performance gap on the low and high complexity tasks. However, relative to human performance these gaps are marginal at best. Broadly, MMICL was the worst-performing model. The expected inverse correlation between complexity and action accuracy was only captured clearly by the GPT-4V and Gemini-Pro-1.0 results.  

In contrast to model performance, human subjects scored much more accurately, with scores ranging from 0.78 to 0.98 across complexities. This model-human gap in performance indicates a significant shortcoming of LMMs on multi-image instruction following tasks. This shortcoming is unlikely due to insufficient feature understanding, as the high single-frame category task performance (Figure \ref{fig:4}b) conflicts with the observed weak category task performances across complexities.

We also analyzed how each model performed on subsets of tasks where single or multiple object properties were involved. Figure \ref{fig:4} b-e shows the average accuracy of all models conditioned on task properties and complexity level. We did not include the object-identity subgroup in the high complexity plot of Figure \ref{fig:4}e as non-boolean response types for object identity tasks are not feasible. The performances on high complexity tasks show a clear ranking between the abilities of GPT-4V and Gemini-pro models on tasks with diverse response options. For almost all models and complexities, location-only tasks posed the most difficulties. This finding confirms previous analyses of GPT-4V, which found that it often struggles to correctly recognize an object's position within an image\footnote{https://blog.roboflow.com/gpt-4v-object-detection} \citep{meta2024}.

To further investigate the response patterns of the LMMs we performed further analyses on a subset of the models seen in Appendix Figures \ref{fig:S6}, \ref{fig:S7}, \ref{fig:S8}, and \ref{fig:S9}. To determine the effects of delay frames on model performance a set of simple delayed-match-to-sample tasks were generated which only differed in difficulty by the number of delay frames. Appendix Figure \ref{fig:S6} shows that for InternLM-XComposer2, MMICL, and GPT-4v, the addition of single or multiple delay frames in a task has little effect on task performance. Next, we looked into how the number of different boolean operators affected open-source model (InternLM-XComposer2 \& MMICL) performance. The Appendix Figure \ref{fig:S7} a-d results show that, generally, across all boolean operators an increase in their abundance within a task leads to worse model performance, which is expected. In Appendix Figure \ref{fig:S8} we examined how the number of stimuli affected task performance across complexities. We expected to see that as the number of stimuli increases, the task performance would decrease. This was found to be the case for Low complexity tasks, which can be seen in Appendix Figure \ref{fig:S8} a. However Medium and High complexity tasks displayed an inverse trend, as seen in Appendix Figure \ref{fig:S9} b and c. Finally, we investigated the exact effect that different required response types had on accuracy for the High complexity benchmark. As Appendix Figure \ref{fig:S9} shows, when tasks required a response that was a non-boolean type word, the models performed significantly worse. 

\section{Conclusion and future directions} 

We introduced iWISDM as a platform for validating multimodal models. As a benchmark, our primary focus is on assessing the ability of LMMs to follow instructions in visual-language decision-making tasks. 
We developed three benchmarks of incremental complexity and evaluated several LMMs and human performance. The gap between LMMs and humans indicates that LMMs still lack key abilities to solve instruction-following tasks. Through a detailed analysis of LMMs' behaviour patterns, we identified diminished spatial recognition ability and decreased performance with increased task complexity.

We believe iWISDM would be an important benchmark that complements existing benchmarks which evaluate LMM capabilities in areas such as commonsense reasoning, numerical computation, or relational inferences \citep{fu2023mme}. Although the current version of iWISDM lacks the capacity to probe these functions, it could potentially cover some of these additional capabilities by adding new operators like Count (to tally specific objects) or Relative (to identify properties like location in relation to other objects). We also believe iWISDM can serve as an important benchmark for continual learning algorithms evaluation. For detailed discussions, please refer to Appendix \ref{appendix_extensible}.

Moreover, the currently used stimulus dataset is derived from publicly accessible sources \citep{chang2015shapenet}, which may pose a risk of data leakage. To address this, we plan to introduce a variety of datasets that users can easily select from. Finally, the compositional nature of iWISDM tasks may also provide an avenue for exploring the failure modes of current LMMs by investigating their specific weaknesses and specifically targeting those during training. We are looking to develop detailed evaluation criteria tailored to iWISDM and to establish an evaluation platform, along with a continuously updated leaderboard, for the purpose of measuring and comparing the performances of models.

\subsubsection*{Acknowledgments}
X.L. was supported by CAMBAM fellowship (2024) and Doctoral Excellence Scholarship, Union Neuroscience et Artificial Intelligence Quebec, UNIQUE (2021-2022). L.G. was supported by Masters Excellence Scholarship, Union Neuroscience et Artificial Intelligence Quebec, UNIQUE (2024). This research was supported by the Healthy-Brains-Healthy-Lives startup supplement grant and the NSERC Discovery grant RGPIN-2021-03035. P.B. was supported by FRQ-S Research Scholars Junior 1 grant 310924, and the William Dawson Scholar award. All analyses were executed using resources provided by the Digital Research Alliance of Canada (Compute Canada) and funding from Canada Foundation for Innovation project number 42730. The funders had no role in study design, data collection and analysis, decision to publish, or preparation of the manuscript.

\bibliography{collas2024_conference}
\bibliographystyle{collas2024_conference}
\newpage
\appendix
\section{Appendix}
\setcounter{figure}{0}
\setcounter{table}{0}
\renewcommand\thefigure{A\arabic{figure}}  
\renewcommand\thetable{A\arabic{table}}

\subsection{Additional Discussion}
\subsubsection{Extensibility of iWISDM}
\label{appendix_extensible}
\begin{itemize}
    \item \textbf{Task operators}. iWISDM task rules are constructed from the interconnection of various building blocks called task operators. Task operators themselves are highly customizable and extensible, allowing users to define new task logic by inheriting the \texttt{Operator} class and overriding the \texttt{get\_expected\_input} function. Our core task operator set is adopted from a prior work \citep{yang2018dataset} and includes \texttt{Get}, \texttt{IsSame}, \texttt{NotSame}, \texttt{And}, \texttt{Or}. 
    \item \textbf{Visual inputs}. iWISDM also allows any natural image set to be used as the stimulus set. In our core build, we use 2D projection of 3D object models from the ShapeNet dataset \citep{chang2015shapenet}, where each stimulus is defined by a parameter vector consisting of \texttt{category}, \texttt{identity}, \texttt{pose angle}, \texttt{location}. We provide a template that allows users to seamlessly import alternative stimulus datasets. 
    \item \textbf{Stimulus properties}. In addition to the visual stimuli themselves, users can define arbitrary object/stimuli properties to be used by the environment during task generation. In our core build of iWISDM, we use \texttt{category}, \texttt{identity}, \texttt{pose angle}, \texttt{location} as the default object properties used during task generation. These properties are attached to each individual object in our stimulus set via an accompanying JSON file. Users can add custom properties by editing the accompanying JSON files for their stimuli of choice. 
\end{itemize}

\subsubsection{iWISDM as a framework for Continual Learning}
\label{appendix_discussion}
Additionally, iWISDM provides a framework for testing and comparing different approaches for continual learning (CL) and multi-task learning. iWISDM can be used to generate any number of tasks with quantifiable similarity which can in turn be used to test CL approaches on their capacity to sequentially learn tasks with different degrees of similarities without forgetting. Unlike most current multi-task learning models tested in fixed environments with relatively homogeneous task structures, iWISDM introduces dynamic task variations. For instance, in computer vision, CL algorithms are typically tested on classifying an increasing number of image classes using datasets such as MNIST \citep{deng2012mnist}, CORe50 \citep{lomonaco2017core50}, or similar datasets. Task-free continual learning approaches \citep{aljundi2019task} exist but are still limited in generating a continual stream of tasks. In contrast, iWISDM allows users to generate a series of tasks with monotonically increasing or decreasing difficulty levels by adjusting complexity parameters in a fully controllable manner.

A robust model should adapt to changing tasks based on prior sequential experiences. Therefore, CL models should be evaluated on their learning speed for incoming tasks, considering quantifiable complexity parameters and confounding factors. Existing CL models often suffer from rapid performance degradation, known as catastrophic forgetting. iWISDM can be used to test the hypothesis that properly ordered learning sequences of incoming tasks can prevent catastrophic forgetting, a concept known as curriculum learning. 

Achieving human-level intelligence requires models capable of out-of-distribution generalization. Compositional generalization is an important ability of intelligent agents and 
thus, it is important to evaluate models on their ability to generalize to compositionally generated scenarios. iWISDM's graph structure allows for the manual construction of novel tasks based on existing task graphs, enabling systematic testing of both logical and temporal compositional generalization abilities. This makes iWISDM an ideal test bed for modularity-based CL algorithms, as it facilitates parallel connections between modules of models and tasks.

iWISDM is also suitable for evaluating meta-learning models in the CL domain. While natural language instructions are provided, other methods of constructing task-specific identifiers, such as one-hot compressed encoding of the task graph using simple graph theory methods, are available. These different methods for representing task-specific identifiers can serve as templates for testing against the learned representations from a meta-learner. Analyzing the similarity across these identified identifiers can further our understanding of the underlying task-solving mechanisms of neural network models.

\subsection{Additional Figures}
\begin{figure}[h]
\centering
\includegraphics[width=1.0\textwidth, scale=0.1]{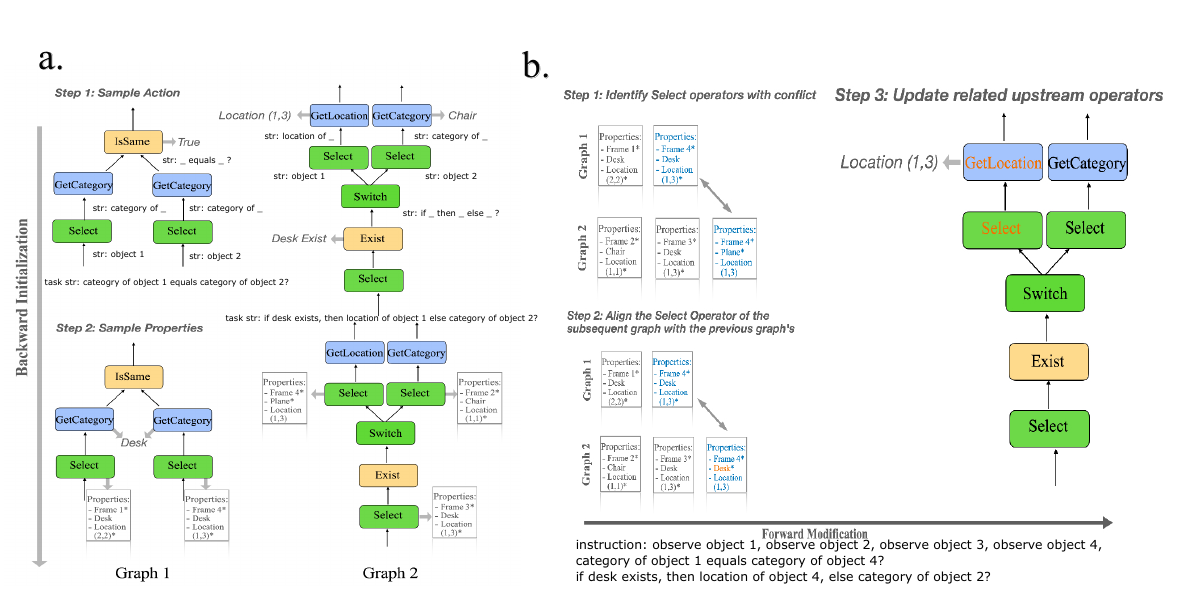}

\caption{Task graph initialization and modification in iWISDM. This figure demonstrates the process iWISDM follows to initialize and modify a set of task graphs. We demonstrate the process using two examples: Graph 1 and Graph 2. \textbf{Backward Initialization a)}: \textit{Step 1}: Identify all operators linked to explicit (e.g. \texttt{IsSame} operator in Graph 1, and \texttt{Get} operators in Graph 2) or implicit responses (where agents decide without a direct response, typically preceding the \texttt{Switch} operator, like the \texttt{Exist} operator in Graph 2), and assign responses to each of these operators, manually balancing output action. \textit{Step 2}: Propagate properties to downstream operators. For example, if the response to \texttt{IsSame} is True, then the children \texttt{GetCategory} operators must have the same output. The \texttt{Select} operator usually receives one property from upstream operators, such as the category Desk in Graph 2's bottom \texttt{Select}. In this example, ``when" and ``where" properties are randomly sampled and marked with asterisks. To illustrate instruction generation, the operator partial strings are shown as ``str: '', and the blanks are filled by its children operators, following the direction of the arrows. However, during temporal composition, backward initialization can create conflicts between graphs, which leads \textbf{Forward Modification Phase b)} to resolve these issues.
\textit{Step 1}: For each graph, gather properties from each \texttt{Select} operator and compare those with matching ``when" to identify conflicts. For instance, a conflict is found in the ``what" property at frame 4; Graph 1 \texttt{Select} assigns ``Desk" while Graph 2's assigns ``Plane". Conflicts can also occur with the ``where" property. We modify the \texttt{Select} in new task graphs to minimize upstream modifications. \textit{Step 2}: Adjust the ``what" property in Graph 2 to ``Desk", aligning with the Select operator in Graph 1. \textit{Step 3}: Post-modification, it may be necessary to update actions or properties of upstream operators. In this scenario, tracing back to the \texttt{GetLocation} operator in Graph 2 shows that no changes are required for the ``where" property, as it remains consistent with the \texttt{Select} operator's assignment.} 
\label{fig:S1}
\end{figure}

\begin{figure}[h]
\centering
\includegraphics[width=1.0\textwidth, scale=0.1]{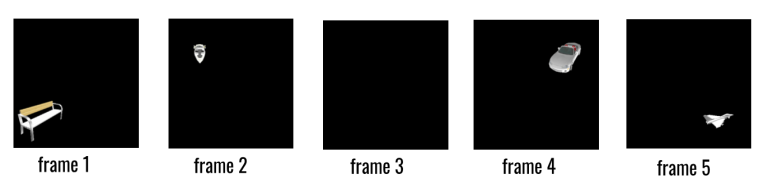}

\caption{An example of low complexity task generated by iWISDM. Instruction: "observe object 1, observe object 2, delay, observe object 3, observe 4, location of object 3 equals location of object 2 or location of object 1 equals location of object 4?"} 
\label{fig:Ss}
\end{figure}

\begin{figure}[h]
\centering
\includegraphics[width=1.0\textwidth, scale=0.1]{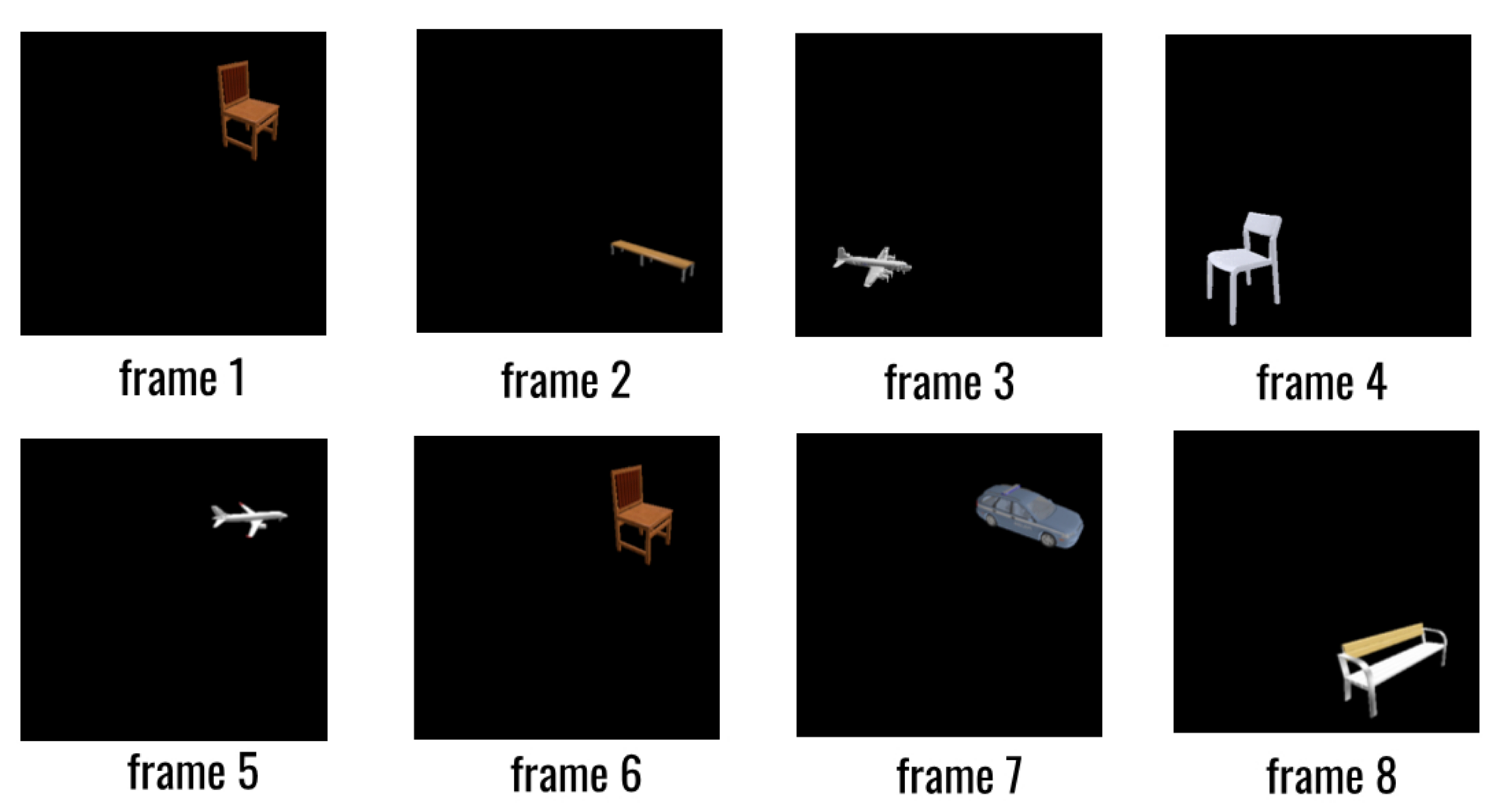}

\caption{An example of medium complexity task generated by iWISDM.Instruction: "observe object 1, observe object 2, observe object 3, observe 4, observe object 5, observe object 6, observe object 7, observe 8, if location of object 7 equals location of object 2 or location of object 8 equals location of object 4, then location of object 3 equals location of object 1? else location of object 6 equals location of object 5?"} 
\label{fig:S3}
\end{figure}

\begin{figure}[h]
\centering
\includegraphics[width=1.0\textwidth, scale=0.1]{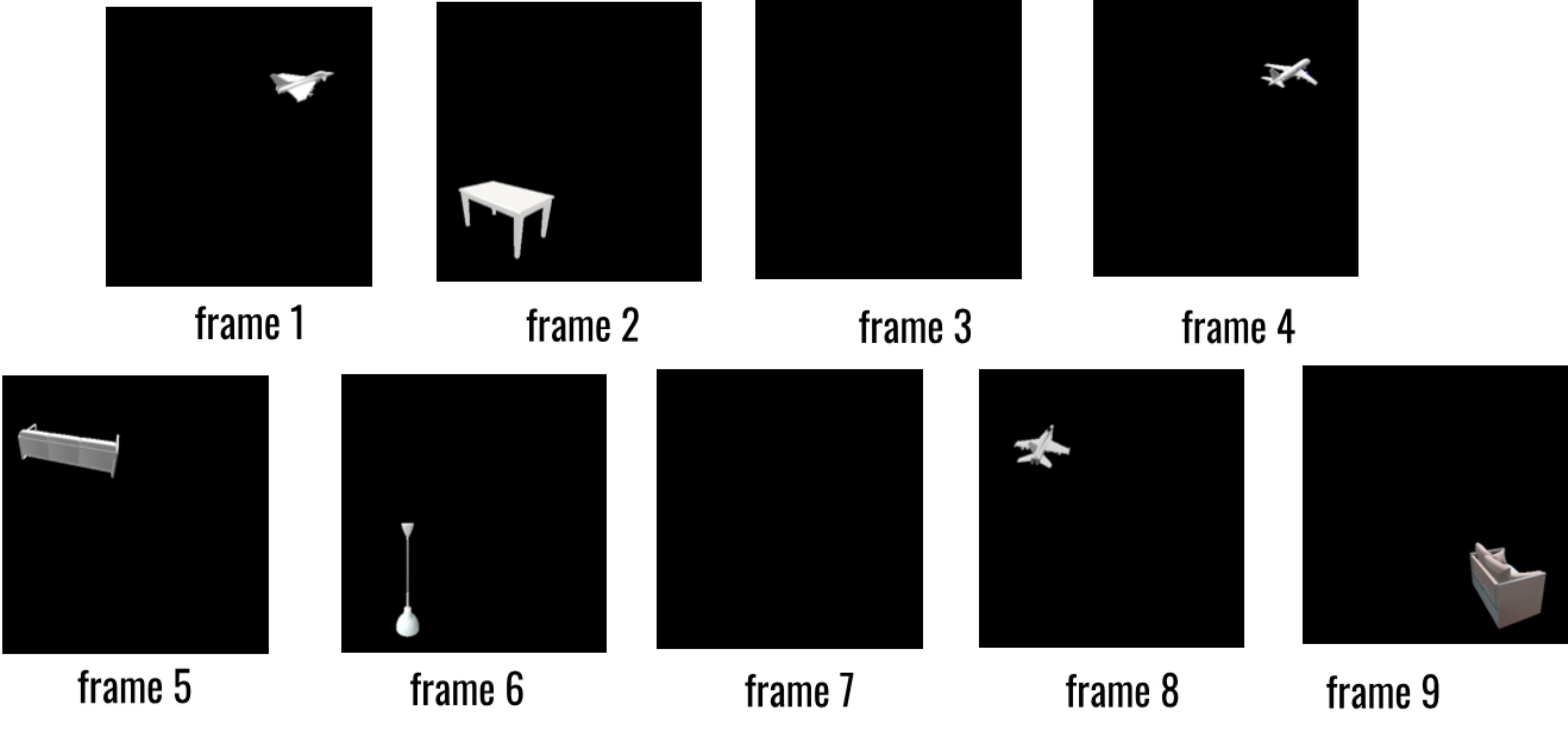}

\caption{An example of high complexity task generated by iWISDM. Instruction: "observe object 1, observe object 2, delay, observe object 3, observe 4, observe object 5, delay, observe object 6, observe object 7, if location of object 3 not equals top right and location of object 2 equals location of object 5, then location of object 7 equals top left and location of object 6 equals location of object 4? else location of object 1?"} 
\label{fig:S4}
\end{figure}

\begin{figure}[h]
\centering
\includegraphics[width=1.0\textwidth, scale=0.1]{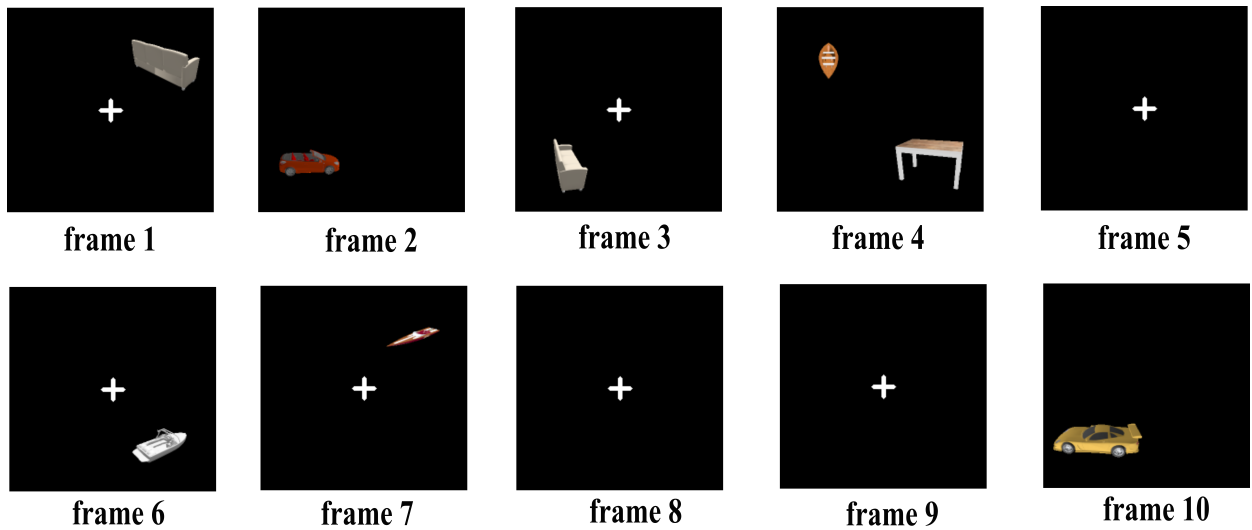}

\caption{An example trial with distractors. Instruction: "observe object 1, observe object 2, category of object 2 not equals category of object 1? observe object 3,observe object 4 with location: top left, category of object 4 equals couches or identity of object 3 equals identity of object 1? delay, observe object 5, observe object 6, delay, observe object 7, identity of object 7 equals identity of object 6 and identity of object 4 equals identity of object 3?"}
\label{fig:S4-distractors}
\end{figure}

\begin{figure}[h]
\centering
\includegraphics[width=0.5\textwidth, scale=0.1]{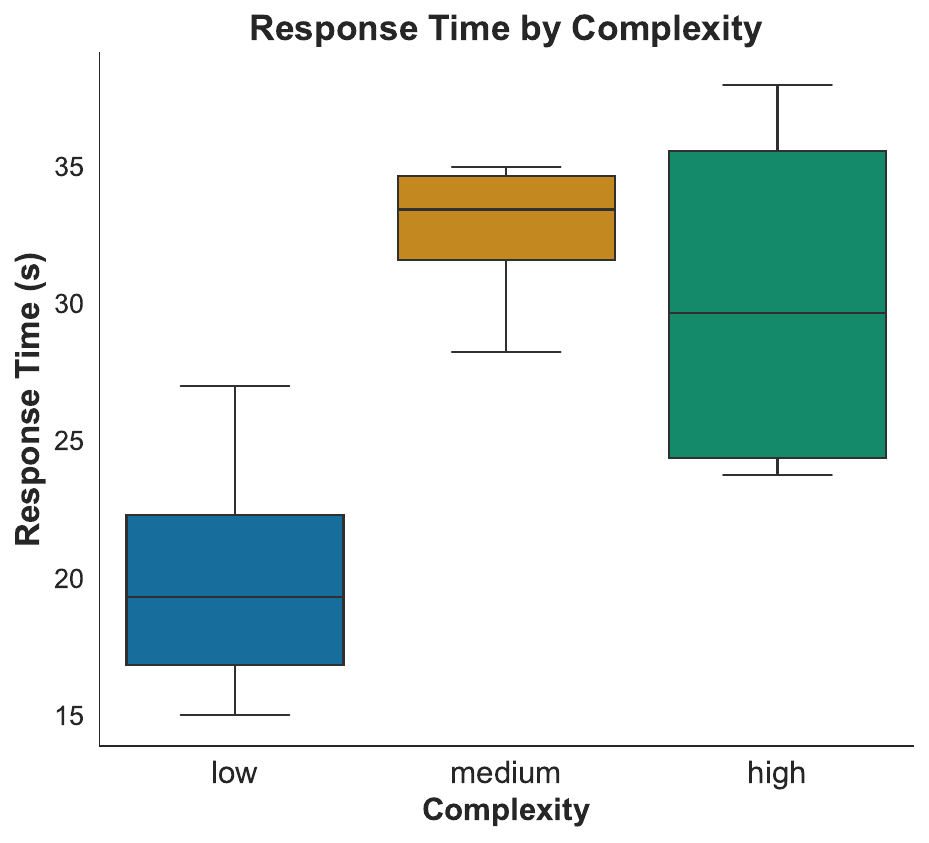}

\caption{Human response times across complexity levels.}
\label{fig:S5}
\end{figure}

\begin{figure}[h]
\centering
\includegraphics[width=0.89\textwidth, scale=1]{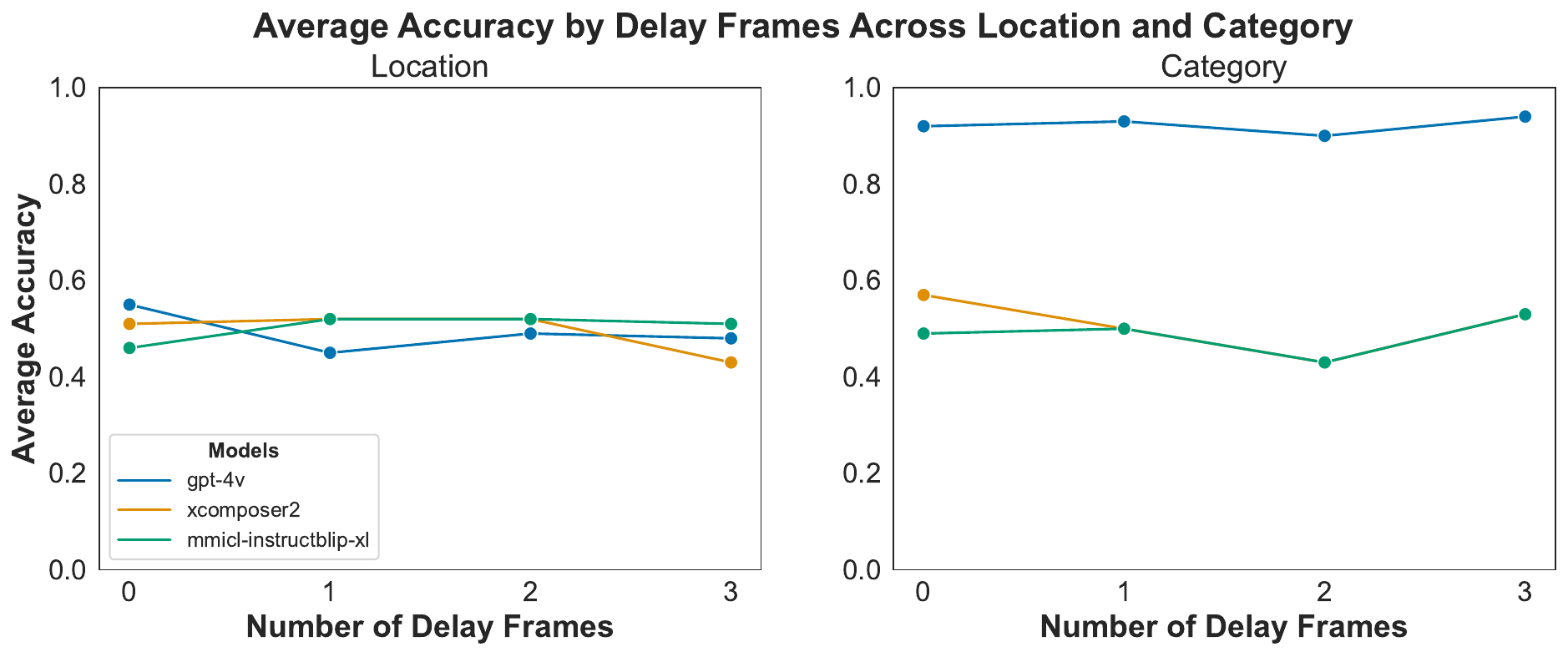}

\caption{Average accuracy of GPT-4v, xcomposer2, and MMICL across a varying number of delay frames.}
\label{fig:S6}
\end{figure}

\begin{figure}[h]
\centering
\includegraphics[width=1\textwidth, scale=1]{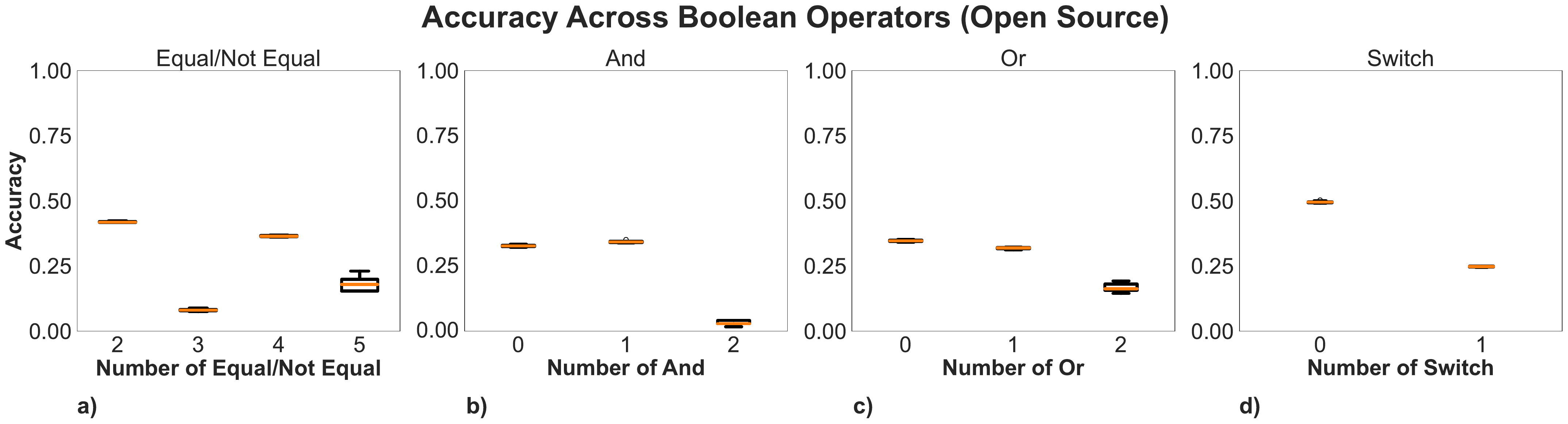}

\caption{\textbf{a)} The performance accuracy of xcomposer2 and MMICL over the number of \textit{Equal/Not Equal} boolean task operators. \textbf{b)} The performance accuracy of xcomposer2 and MMICL over the number of \textit{And} boolean task operators. \textbf{c)} The performance accuracy of xcomposer2 and MMICL over the number of \textit{Or} boolean task operators. \textbf{d)} The performance accuracy of xcomposer2 and MMICL over the number of \textit{Switch} task operators.}
\label{fig:S7}
\end{figure}

\begin{figure}[h]
\centering
\includegraphics[width=0.75\textwidth, scale=1]{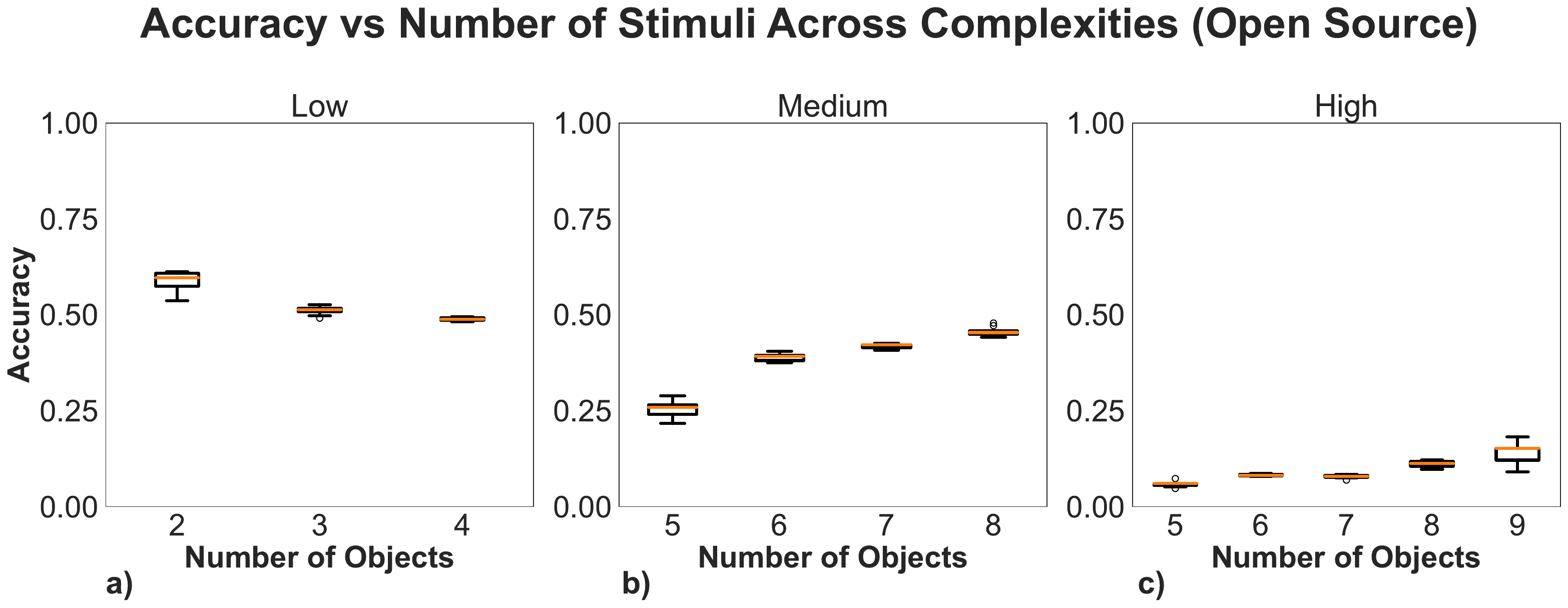}

\caption{\textbf{a)} The performance accuracy of xcomposer2 and MMICL on low complexity tasks with varying numbers of object stimuli. \textbf{b)} The performance accuracy of xcomposer2 and MMICL on medium complexity tasks with varying numbers of object stimuli. \textbf{c)} The performance accuracy of xcomposer2 and MMICL on high complexity tasks with varying numbers of object stimuli.}
\label{fig:S8}
\end{figure}

\begin{figure}[h]
\centering
\includegraphics[width=0.68\textwidth, scale=1]{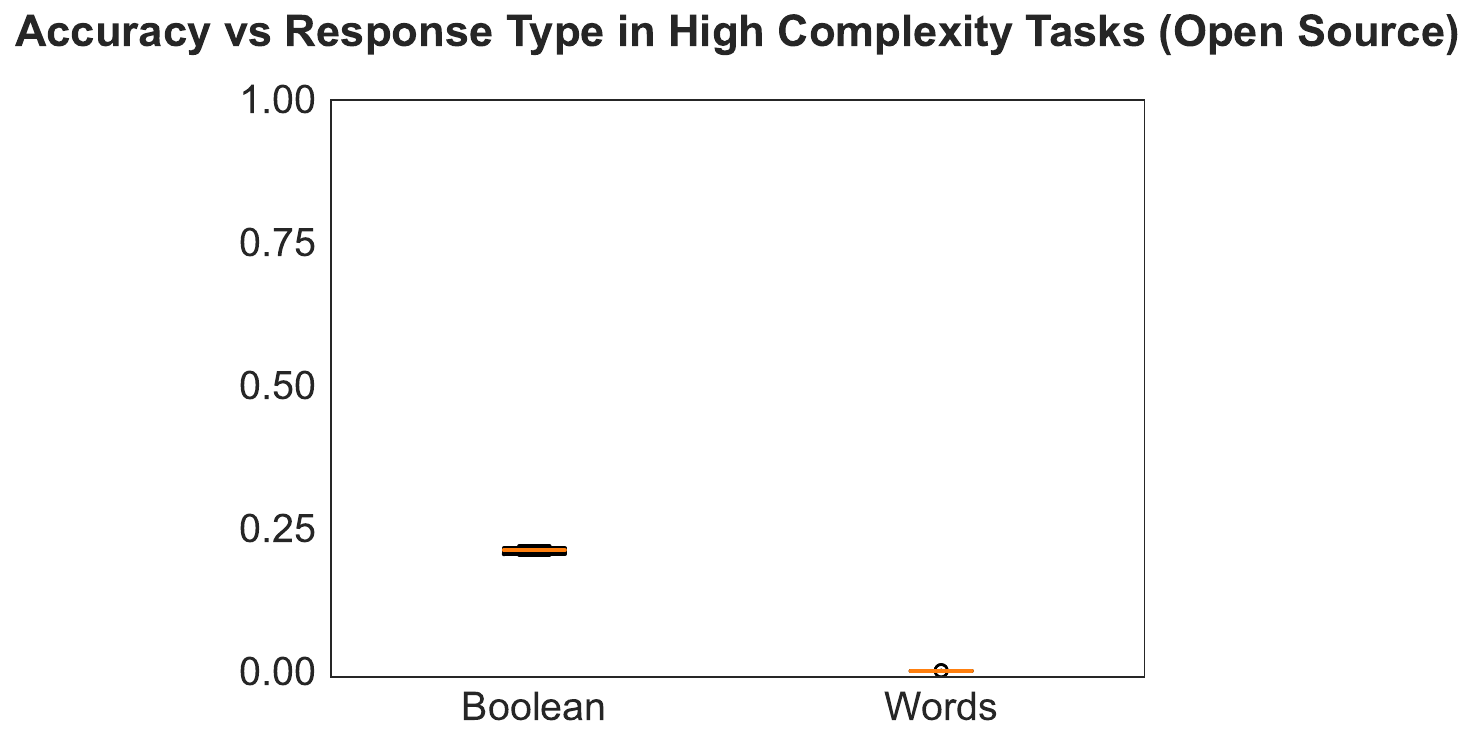}

\caption{Accuracy of xcomposer2 and MMICL across response types for high complexity tasks.}
\label{fig:S9}
\end{figure}

\clearpage
\subsection{Model evaluation prompts}
\label{fig:prompts}
\subsubsection{GPT-4V, Claude-3, \& Gemini-pro Low/Medium Complexity (all properties + examples included)} 
\texttt{In this task we will show you a series of frame images. Each frame will either be blank (delay frame) or contain a 3D object.The objects within the task will ALWAYS be from one of 8 categories: benches, boats, cars, chairs, couches, lighting, planes, and tables. For each of these 8 categories, there are 8 unique objects that could be used in the task. Any object which is sampled will be displayed as an image taken from a random viewing angle. The objects will be placed in one of four locations: top left, top right, bottom left, and bottom right.}

\texttt{A written instruction will be provided. Your goal is to follow the instructions and answer the question contained in the instruction. Answers will ALWAYS be one of the following: true, false .}

\texttt{Here is a simple example of the task ... }

\texttt{Task instruction: "observe object 1, observe object 2, location of object 1 not equal location: bottom left ?"} 

\texttt{Here are the corresponding frames ...}

\texttt{Answer: false.
This is because the location of object 1 IS in the bottom left location. }

\texttt{Here is a simple example of the task...} 

\texttt{Task instruction: "observe object 1, delay, observe object 2, category of object 1 equals category of object 2 ?"}

\texttt{Here are the corresponding frames ...}

\texttt{Answer: true.
This is because the category of object 1 (lighting) IS equal to the category of object 2 (lighting). }

\texttt{Here is a simple example of the task... }

\texttt{Task instruction: "observe object 1, observe object 2, identity of object 1 equals identity of object 2 ?" }

\texttt{Here are the corresponding frames ...}

\texttt{Answer: true.}
\texttt{This is because object 1 (a white table) IS identical to object 2 (the same white table). }

\texttt{Now please solve the following new task...}

\texttt{Task instruction: "observe object 1, observe object 2, delay, observe object 3, observe object 4, category of object 2 not equal category of object 3 or category of object 1 equals category of object 4 ?"}

\texttt{Here are the corresponding frames ...}

\texttt{What is the correct answer to this task? (respond EXACTLY and ONLY with one of the following answers: true, false). Provide your answer here:}

\subsubsection{GPT-4V Single-Frame Complexity (location property + no examples included)} 
\texttt{In this task we will show you an image. Each image will contain a 3D object.The objects within the task will ALWAYS be from one of 8 categories: benches, boats, cars, chairs, couches,lighting, planes, and tables. For each of these 8 categories there are 8 unique objects that could be used in the task.Any object which is sampled will be displayed as an image taken from a random viewing angle. The object will be placed in one of four locations: top left, top right, bottom left, and bottom right.}

\texttt{A written instruction will be provided. Your goal is to follow the instructions and answer the question contained in the instruction. Answers will ALWAYS be one of the following: true, false.}

\texttt{Now please solve the following new task...}

\texttt{Task instruction: "observe object 1, category of object 1 not equals planes?"}

\texttt{Here are the corresponding frames ...}

\texttt{What is the correct answer to this task? (respond EXACTLY and ONLY with one of the following answers: true, false). Provide your answer here:}

\subsubsection{GPT-4V, Claude-3, \& Gemini-pro High Complexity (all properties + examples included)} 
\texttt{In this task we will show you a series of frame images. Each frame will either be blank (delay frame) or contain a 3D object.The objects within the task will ALWAYS be from one of 8 categories: benches, boats, cars, chairs, couches,lighting, planes, and tables. For each of these 8 categories there are 8 unique objects that could be used in the task.Any object which is sampled will be displayed as an image taken from a random viewing angle. The objects will be placed in one of four locations: top left, top right, bottom left, and bottom right.}

\texttt{A written instruction will be provided. Your goal is to follow the instructions and answer the question contained in the instruction. Answers will ALWAYS be one of the following: true, false, bottom right, bottom left, top left, top right, benches, boats, cars, chairs, couches, lighting, planes, tables .}

\texttt{Here is an example of the task...}

\texttt{Task instruction: "observe object 1, observe object 2, location of object 1 not equal location: bottom left ?"}

\texttt{Here are the corresponding frames ...}

\texttt{The correct answer: bottom right.
 This is because object 2 is located in the bottom right.}

\texttt{Here is a simple example of the task... }

\texttt{Task instruction: "observe object 1, delay, observe object 2, category of object 1 equals category of object 2 ?" }

\texttt{Here are the corresponding frames ...}

\texttt{Answer: lighting.
 This is because object 1 (a lamp) belongs to the category of lighting. }

\texttt{Here is a simple example of the task... }

\texttt{Task instruction: "observe object 1, observe object 2, identity of object 1 equals identity of object 2?" }

\texttt{ Here are the corresponding frames ...}

\texttt{Answer: true.
 This is because object 1 (a white table) IS identical to object 2 (the same white table). }

\texttt{Now please solve the following new task...}

 \texttt{Task instruction: "observe object 1, observe object 2, delay, observe object 3, observe object 4, observe object 5, if location of object 5 not equal location of object 2 , then location of object 1? else category of object 4 not equal tables or category of object 3 not equal couches?"}

 \texttt{Here are the corresponding frames ...}

\texttt{What is the correct answer to this task? (respond EXACTLY and ONLY with one of the following answers: true, false, bottom right, bottom left, top left, top right, benches, boats, cars, chairs, couches, lighting, planes, tables). Provide your answer here: }

\subsubsection{InternLM-XComposer2 \& MMICL Low Complexity (all properties included + examples excluded)} 
\texttt{In this task we will show you a series of frame images. Each frame will either be blank (delay frame) or contain a 3D object.The objects within the task will ALWAYS be from one of 8 categories: benches, boats, cars, chairs, couches,lighting, planes, and tables. For each of these 8 categories there are 8 unique objects that could be used in the task.Any object which is sampled will be displayed as an image taken from a random viewing angle. The objects will be placed in one of four locations: top left, top right, bottom left, and bottom right. }

\texttt{A written instruction will be provided. Your goal is to follow the instructions and answer the question contained in the instruction. Answers will ALWAYS be one of the following: true, false .}

\texttt{Please solve the following task...}

\texttt{Task instruction: "observe object 1, delay, observe object 2, observe object 3, observe object 4, location of object 1 equals location of object 2 and category of object 3 equals category of object 4?" }

\texttt{Here are the corresponding frames ...
 <ImageHere> <ImageHere> <ImageHere> <ImageHere> <ImageHere> <ImageHere>
What is the correct answer to this task? (respond EXACTLY and ONLY with one of the following answers: true, false). Provide your answer here: }

\subsubsection{InternLM-XComposer2 High Complexity (all properties included + examples excluded)} 
\texttt{In this task we will show you a series of frame images. Each frame will either be blank (delay frame) or contain a 3D object.The objects within the task will ALWAYS be from one of 8 categories: benches, boats, cars, chairs, couches,lighting, planes, and tables. For each of these 8 categories there are 8 unique objects that could be used in the task.Any object which is sampled will be displayed as an image taken from a random viewing angle. The objects will be placed in one of four locations: top left, top right, bottom left, and bottom right.} 

\texttt{A written instruction will be provided. Your goal is to follow the instructions and answer the question contained in the instruction. Answers will ALWAYS be one of the following: true, false, bottom right, bottom left, top left, top right, benches, boats, cars, chairs, couches, lighting, planes, tables .}

\texttt{Please solve the following task...}

\texttt{Task instruction: "observe object 1, observe object 2, observe object 3, observe object 4, delay, observe object 5, observe object 6, delay, observe object 7, if location of object 6 not equal location of object 2 or category of object 3 not equal category of object 4, then location of object 7 equals location of object 5? else category of object 1 ?" }

\texttt{Here are the corresponding frames ...
 <ImageHere> <ImageHere> <ImageHere> <ImageHere> <ImageHere> <ImageHere> <ImageHere> <ImageHere> <ImageHere>
What is the correct answer to this task? (respond EXACTLY and ONLY with one of the following answers: true, false, bottom right, bottom left, top left, top right, benches, boats, cars, chairs, couches, lighting, planes, tables). Provide your answer here: }

\newpage
\begin{table}[h]
\centering
    \caption{Benchmark details for each level of complexity}
    \label{fig:complexity-table}
    \begin{tabular}{l|l|l|l|l|l}
    \textbf{Complexity} & \textbf{\begin{tabular}[c]{@{}l@{}}\# of allowed \\ and/or\\ operators \\ in task\end{tabular}} & \textbf{\begin{tabular}[c]{@{}l@{}}\# of switch\\ operators\end{tabular}} & \textbf{\begin{tabular}[c]{@{}l@{}}\# of trial \\ frames\end{tabular}} & \textbf{\begin{tabular}[c]{@{}l@{}}root\\ operators\end{tabular}}                              & \textbf{\begin{tabular}[c]{@{}l@{}}boolean \\ operators\end{tabular}} \\ \hline
    Low                 & 1                                                                                                & 0                                                                         & 6                                                                      & \begin{tabular}[c]{@{}l@{}}IsSame, And, \\ Or, NotSame\end{tabular}                            & \begin{tabular}[c]{@{}l@{}}IsSame, And, \\ Or, NotSame\end{tabular}   \\ \hline
    Medium              & 1                                                                                                & 1                                                                         & 8                                                                      & \begin{tabular}[c]{@{}l@{}}IsSame, And, \\ Or, NotSame\end{tabular}                            & \begin{tabular}[c]{@{}l@{}}IsSame, And, \\ Or, NotSame\end{tabular}   \\ \hline
    High                & 1-2                                                                                              & 1                                                                         & 9                                                                      & \begin{tabular}[c]{@{}l@{}}IsSame, And, \\ Or, NotSame, \\ GetLoc, \\ GetCategory\end{tabular} & \begin{tabular}[c]{@{}l@{}}IsSame, And, \\ Or, NotSame\end{tabular}  
    \end{tabular}

\end{table}

\end{document}